\documentclass[10pt,twocolumn,letterpaper]{article}

\usepackage{iccv}
\usepackage{times}
\usepackage{epsfig}
\usepackage{graphicx}
\usepackage{amsmath}
\usepackage{amssymb}
\usepackage{url}
\usepackage{soul}
\usepackage{color}
\usepackage{subcaption}
\usepackage{multirow}
\usepackage[multiple]{footmisc}
\usepackage{tabularx,ragged2e,booktabs}
\usepackage{pbox}

\usepackage[breaklinks=true,bookmarks=false]{hyperref}

\iccvfinalcopy 

\def\iccvPaperID{008} 

\begin{document}

\title{Improved Speech Reconstruction from Silent Video}

\author{Ariel Ephrat\thanks{indicates equal contribution} ~~~~~~~~~ Tavi Halperin\footnotemark[1] ~~~~~~~~~ Shmuel Peleg\\
The Hebrew University of Jerusalem\\
Jerusalem, Israel\\
}

\maketitle

\begin{abstract}
Speechreading is the task of inferring phonetic information from visually observed articulatory facial movements, and is a notoriously difficult task for humans to perform. In this paper we present an end-to-end model based on a convolutional neural network (CNN) for generating an intelligible and natural-sounding acoustic speech signal from silent video frames of a speaking person.
We train our model on speakers from the GRID and TCD-TIMIT datasets, and evaluate the quality and intelligibility of reconstructed speech using common objective measurements. We show that speech predictions from the proposed model attain scores which indicate significantly improved quality over existing models. In addition, we show promising results towards reconstructing speech from an unconstrained dictionary.
\end{abstract}

\begin{figure}[t]
   \centering
	\includegraphics[width=0.65\linewidth]{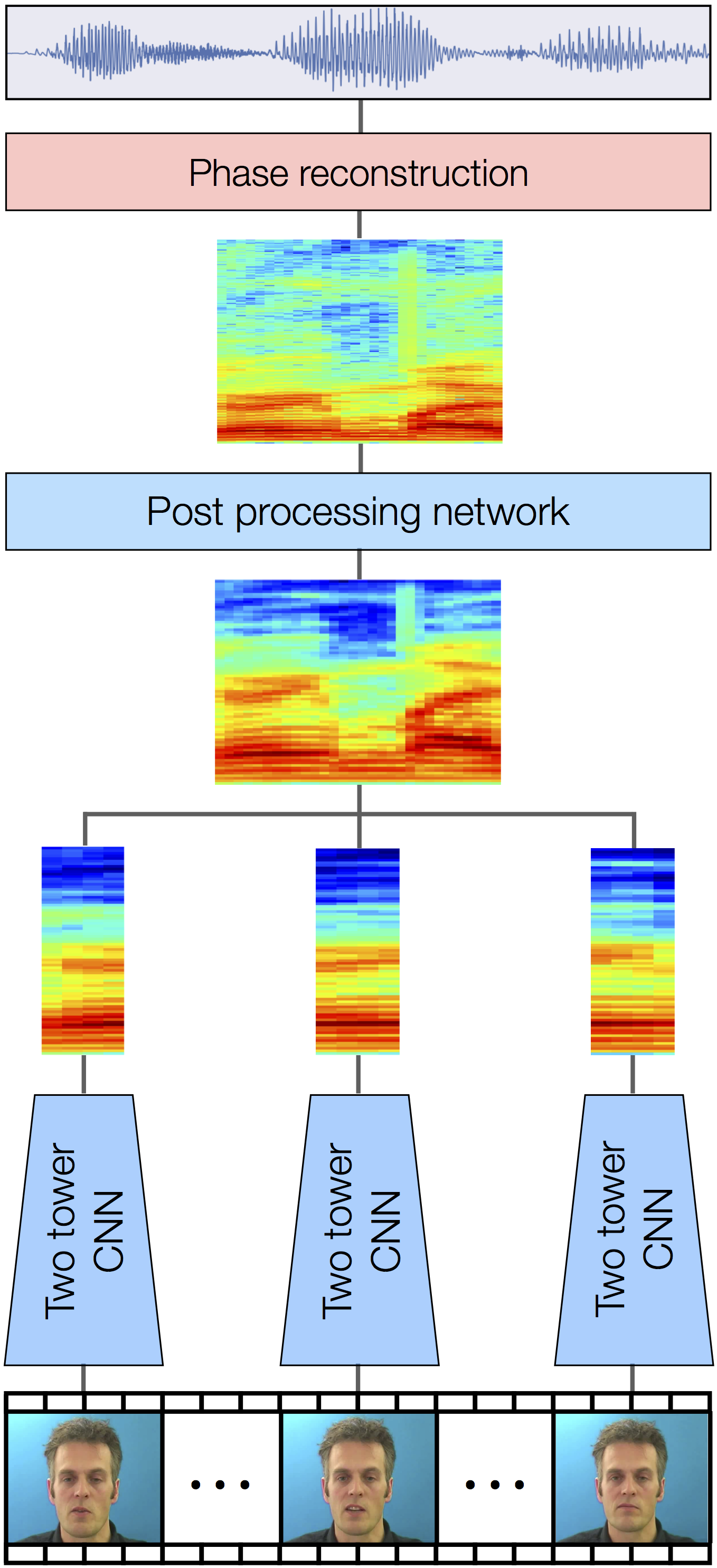}
   \caption{Proposed two-tower CNN-based model converts silent frames of a speaking person to intelligible and natural sounding speech. The CNN generates multiple mel-scale spectrograms which are converted by a post-processing network to a longer range linear-scale spectrogram. The frames-to-spectrogram model is trained end-to-end. Phase reconstruction transforms the long-range spectrogram into waveform.}
   \label{fig:overview}
\end{figure}

\section{Introduction}
\label{sec:intro}

Human speech is inherently an articulatory-to-auditory mapping in which mouth, vocal tract and facial movements produce an audible acoustic signal containing phonetic units of speech (phonemes) which together comprise words and sentences. \emph{Speechreading} (commonly called \emph{lipreading}) is the task of inferring phonetic information from these facial movements by visually observing them. Considering the fact that speech is the primary method of human communication, people who are deaf or have a hearing loss find that speechreading can help overcome many of the barriers when communicating with others \cite{burnham2013hearing}. However, since  several phonemes often correspond to a single viseme (visual unit of speech), it is a notoriously difficult task for humans to perform.

We believe that machine speechreading may be best approached using the same form of articulatory-to-acoustic mapping that creates the natural audio signal, even though not all relevant information is available visually (e.g. vocal chord and most tongue movement). In addition to the perceptual sense this approach makes, modeling the task as an acoustic regression problem has many advantages over the visual-to-textual or classification modeling: $(i)$ Acoustic speech signal contains information which is often difficult or impossible to express in text, such as emotion, prosody and word emphasis; $(ii)$ This form of cross-modal regression, in which one modality is used to generate another modality, can be trained using ``natural supervision'' \cite{owens2015visually} which leverages the natural synchronization in a video of a talking person. Recorded video frames and recorded sound do not require any segmentation or labeling; $(iii)$ By regressing very short units of speech, we can learn to reconstruct words comprised of these units which were not ``seen'' during training. While this is also possible by classifying short visual units into their corresponding phonemes or characters, in practice generating labeled training data for this task is difficult.

Several applications come to mind for automatic video-to-speech systems: Enabling videoconferencing from within a noisy environment; facilitating conversation at a party with loud music between people having wearable cameras and earpieces; maybe even using surveillance video as a long-range listening device. In another paper we have successfully used the generated sound, together with the original noisy sound, for speech enhancement and separation \cite{gabbay2017seeing}.

Our technical approach builds on recent work in neural networks for speechreading and speech synthesis, which we extend to the problem of generating natural sounding speech from silent video frames of a speaking person. To the best of our knowledge, there has been relatively little work for reconstructing high quality speech using an unconstrained dictionary. Our work is also closely related to efforts to extract textual information from a video of a person speaking, i.e. the visual-to-textual problem, as the output of our model can potentially also be used as input to a speech-to-text model.

In this paper, we: (1) present and compare multiple CNN-based encoder-decoder models that predict the speech audio signal of a silent video of a person speaking, and significantly improve both intelligibility and quality of speech reconstructions of existing models; (2) show significant progress towards reconstructing words from an unconstrained dictionary.

\begin{figure*}[t]
   \centering
	\includegraphics[width=\linewidth]{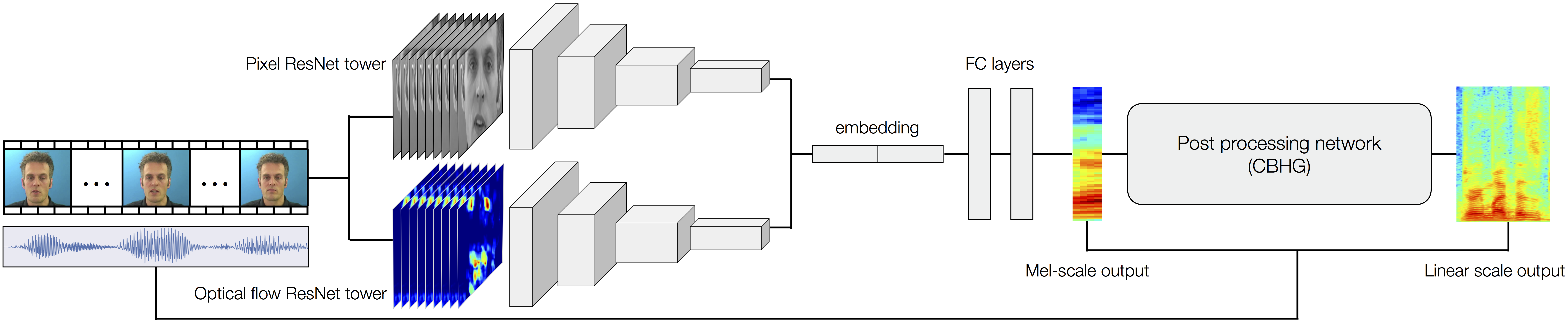}
   \caption{This figure illustrates our model's two-tower CNN-based encoder-decoder architecture. First, a face is detected in the silent input video. The face undergoes in-plane registration, cropping and resizing. One tower inputs grayscale face images, and the other inputs optical flow computed between frames. An embedding is created by concatenating the outputs of each tower. It is subsequently fed into the decoder which consists of fully connected layers which output mel-scale spectrogram, and a post-processing network which outputs linear-scale spectrogram.}
   \label{fig:enc_dec_fig}
\end{figure*}

\section{Related work}
\label{sec:related_work}

Much work has been done in the area of automating speechreading by computers \cite{petajan1984automatic, matthews2002extraction, zhou2014review}. There are two main approaches to this task. The first, and the one most widely attempted in the past, consists of modeling speechreading as a visual-to-textual mapping. In this approach, the input video is manually segmented into short clips which contain either whole words from a predefined dictionary, or parts of words comprising phonemes, visemes \cite{bear2016decoding} or characters. Then, visual features are extracted from the frames and fed to a classifier. Assael \etal \cite{assaellipnet}, Chung \etal \cite{chung2016lip} and others \cite{wand2016lipreading, stafylakis2017combining,petridis2017end} have all recently showed state-of-the-art word and sentence-level classification results using neural network-based models.

The second approach, and the one used in this work, is to model speechreading as a visual-to-acoustic mapping problem in which the ``label" of each short video segment is a corresponding feature vector representing the audio signal. Kello and Plaut \cite{kello2004neural} and Hueber and Bailly \cite{hueber2016} attempted this approach using various sensors to record mouth movements, whereas Cornu and Milner \cite{cornu2017} used active appearance model (AAM) visual features as input to a recurrent neural network.

Our approach is closely related to recent speaker-dependent video-to-speech work by Ephrat and Peleg \cite{vid2speech}, in which a convolutional neural network (CNN) is trained to map raw pixels of a speaker's face directly to audio features, which are subsequently converted into intelligible waveform. The differences between our approach and the one taken by \cite{vid2speech} can by broken down into two parts: improvement of the encoder and redesign of the decoder.

The goal of our encoder modification is to improve analysis of facial movements, and consists of a preprocessing step which registers the face to a canonical pose, the addition of an optical flow branch, and swapping the VGG-based architecture with a ResNet based one.
Our decoder is designed to remedy a major flaw in \cite{vid2speech}, namely the unnatural sound of the reconstructed speech it produces.  To this end, we use the sound representation and post-processing network of \cite{tacotron}, which introduces longer-range dependency into the final speech reconstruction, resulting in smoother, higher quality speech. Section \ref{sec:arch} expounds on the above differences, and Section \ref{ssec:tasks} contains a comparison of the results of \cite{vid2speech} to ours.

Our work also builds upon recent work in neural sound synthesis using predicted spectrogram magnitude, including Tacotron (Wang \etal) \cite{tacotron} for speech synthesis, and the baseline model of NSynth (Engel \etal) for music synthesis. While Tacotron focuses on building a single-speaker text-to-speech (TTS) system, our paper focuses on building a single-speaker video-to-speech system.

This work complements and improves upon previous efforts in a number of ways: Firstly, we explore how to better analyze the visual input, i.e. silent video frames, in order to produce an encoding which can be subsequently decoded into speech features. Secondly, while prior work has predicted only the output corresponding to a single video frame, we jointly generate audio features for a sequence of frames, as depicted in Figure \ref{fig:overview}, which improves the smoothness of the resulting audio. Thirdly, \cite{vid2speech} focused on maximizing intelligibility at the expense of natural sounding speech on a relatively limited-vocabulary dataset. We aim to overcome the challenges of the more complex TCD-TIMIT \cite{tcd-timit} dataset, while optimizing for both intelligibility and natural-sounding speech.

\section{Data representation}
\label{sec:representation}

\subsection{Visual representation}
\label{ssec:visual}

Our goal is to reconstruct a single audio representation vector $S_i$ which corresponds to the duration of a single video frame $I_i$. However, instantaneous lip movements such as those in isolated video frames can be significantly disambiguated by using a temporal neighborhood as context. Therefore, the encoder module of our model takes two inputs: a clip of $K$ consecutive grayscale video frames, and a ``clip'' of $(K-1)$ consecutive dense optical flow fields corresponding to the motion in $(u,v)$ directions for pixels of consecutive grayscale frames.

Each clip is registered to a canonical frame of reference. We start by detecting $5$ facial points (two eyes, nose, and two tips of the mouth). We use the points on the eyes to compute a similarity transform alignment between each frame and the central frame of the clip. Following \cite{vid2speech}, we then crop the speaker's full face to a size of $H \times W$ pixels, and we use the entire face region rather than using only the region of the mouth. This results in an input volume of size $H \times W \times K$ scalars. The second input, dense optical flow, adds an additional volume of $H \times W \times (K-1)\times2$ scalar inputs. It has been proven that optical flow can improve the performance of neural networks when combined with raw pixel values for a variety of applications \cite{Simonyan2014nips,Feich2016cvpr}, and has even been successfully used as a stand-alone network input \cite{poleg2016compact}. Optical flow is positively influential in this case as well, as we show later.

\subsection{Speech representation}
\label{ssec:speech}

The challenge of finding a suitable representation for an acoustic speech signal which can be estimated by a neural network on one hand, and synthesized back into intelligible audio on the other, is not trivial. Use of raw waveform as network output was ruled out for lack of a suitable loss function with which to train the network.

Line Spectrum Pairs (LSP) \cite{lsp} are a representation of Linear Predictive Coding (LPC) coefficients which are more stable and robust to quantization and small deviations. LSPs are therefore useful for speech coding and transmission over a channel, and were used by \cite{vid2speech} as output from their video-to-speech model. However, without the original excitation, the reconstruction using unvoiced excitation (random noise) results in somewhat intelligible, albeit robotic and unnatural sounding speech. 

Given the above, we sought to use a representation which retains speech information vital for an accurate reconstruction into waveform. We experiment with both spectrogram magnitude and a reduced dimensionality mel-scale spectrogram as our regression target, which can subsequently be transformed back into waveform by using a phase reconstruction algorithm such as Giffin-Lim \cite{griffin1984signal}.

\section{Model architecture}
\label{sec:arch}
At a high-level, as shown in Figure \ref{fig:enc_dec_fig}, our model is a comprised of an encoder-decoder architecture which takes silent video frames of a speaking person as input, and outputs a sound representation corresponding to the speech spoken during the duration of the input.

It is important to note that our proposed approach is speaker-dependent, i.e. a new model needs to be trained for each new speaker. Achieving speaker-independent speech reconstruction is a non-trivial task, and is out of the scope of this work.

\subsection{Encoder}
\label{ssec:encoder}
The encoder module of our model consists of a dual-tower Residual neural network (ResNet\cite{he2016deep}) which takes the aforementioned video clip and its optical flow as inputs and encodes them into a latent vector representing the visual features. Each of the inputs is processed with a column of residual block stacks. Each tower comprises ten consecutive $conv1{\times}1-Batch Normalization-Relu-conv3{\times}3-Batch Normalization-Relu-conv1{\times}1-Batch Normalization-Add-Relu$ blocks consisting of $128-128-128-256-256-256-256-512-512-512-512$ kernels. Following the last layer, a global spatial average ($5{\times}4$) is performed, after which the two towers are concatenated into one $1024$-neuron layer which is essentially a latent representation of our visual features.

\subsection{Decoder and post-processing}
\label{ssec:decoder}
The latent vector is fed into a series of two fully connected layers with $1024$ neurons each. The last layer of our CNN is of size $n \times l$, where $l$ is the number of audio windows predicted given a single video clip, and $n$ corresponds to the size of the sound representation vectors we wish to predict. For example, an output of $80$ coefficients of Mel frequency with $hop size = window size / 4$ and window size of $500$, results in an $80 \times 4 = 320$ dimensional output vector. The output of our CNN is fed into the post-processing network used by \cite{tacotron}, consisting of one CBHG module, which is described as a powerful module for extracting representations from sequences. The CBHG module comprises several convolutional, Highway \cite{highway} and Bidirectional GRU \cite{gru} layers whose goal is to take several consecutive sound representations as input and output a higher temporal resolution version. The input clips are then packed in mini-batches of $T$ consecutive samples. As in the implementation of \cite{tacotron}, the post processing network takes these consecutive mel-scale spectrogram vectors as input, and outputs a $T$ consecutive linear-scale spectrogram. The entire model is trained end-to-end with a two-term loss, one on the decoder output and one on the output of the post-processing network. Although the model is trained end-to-end, we keep all convolutional layers of the CNN frozen during training of the second network.

\begin{figure*}
\centering
\begin{subfigure}[b]{0.475\textwidth}
\centering
\includegraphics[width=\textwidth]{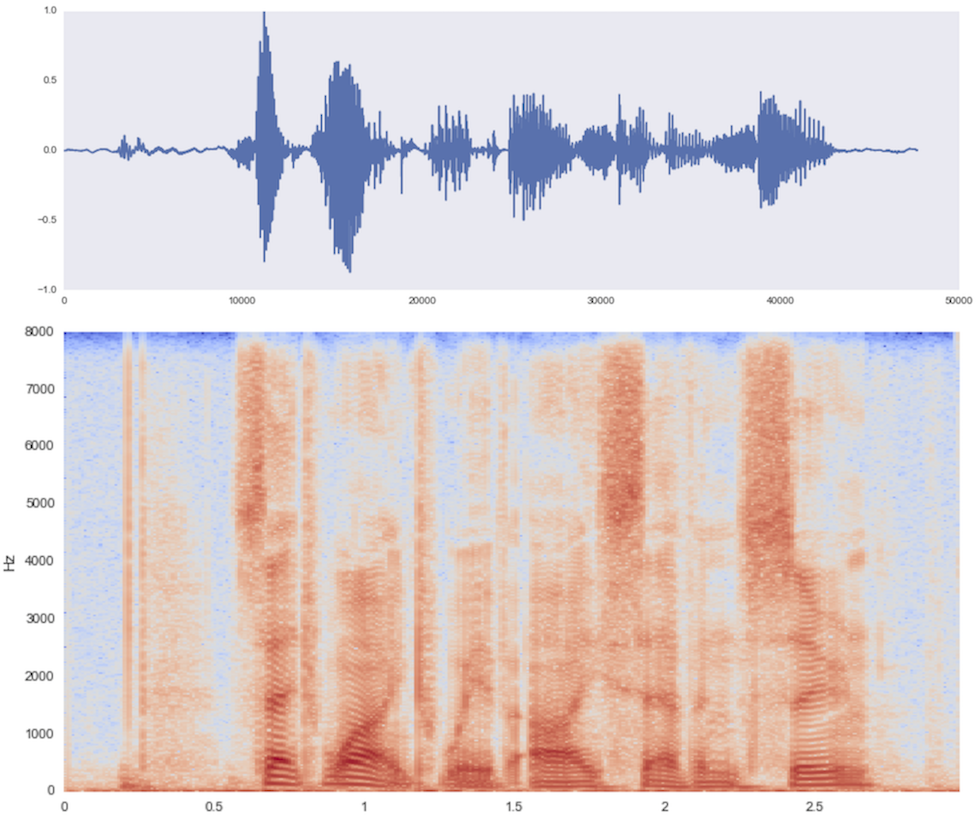}
\caption[]%
{{\small Original}}    
\end{subfigure}
\hfill
\begin{subfigure}[b]{0.475\textwidth}  
\centering 
\includegraphics[width=\textwidth]{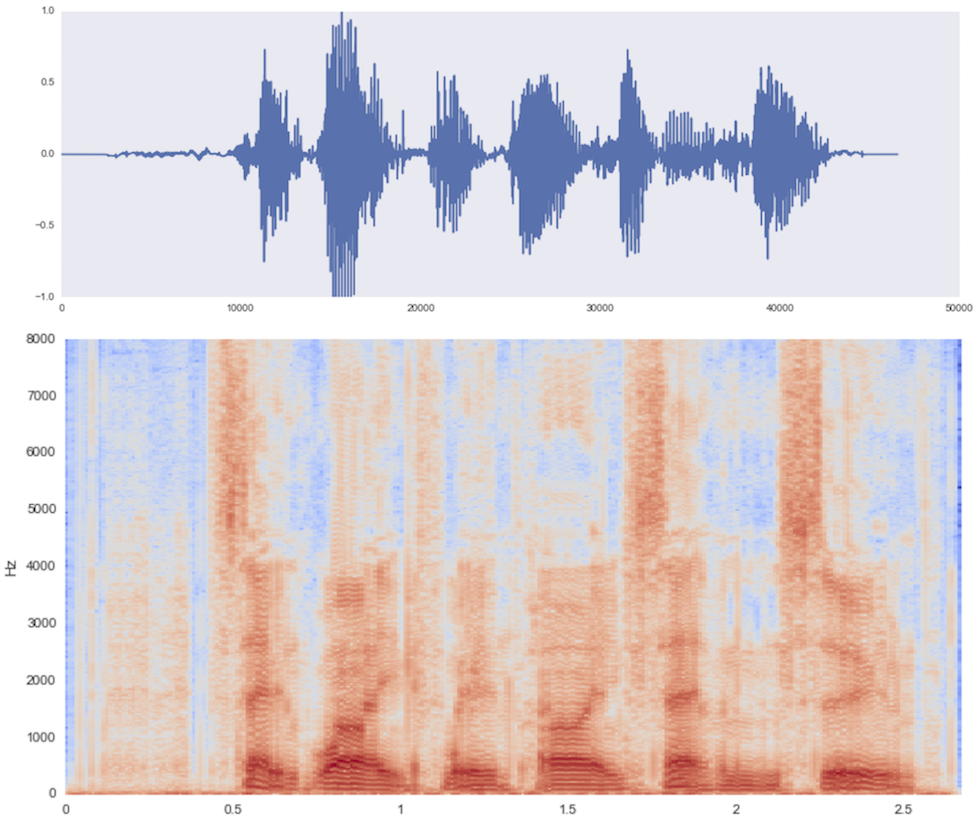}
\caption[]%
{{\small Ours}}    
\end{subfigure}
\vskip\baselineskip
\begin{subfigure}[b]{0.475\textwidth}   
\centering 
\includegraphics[width=\textwidth]{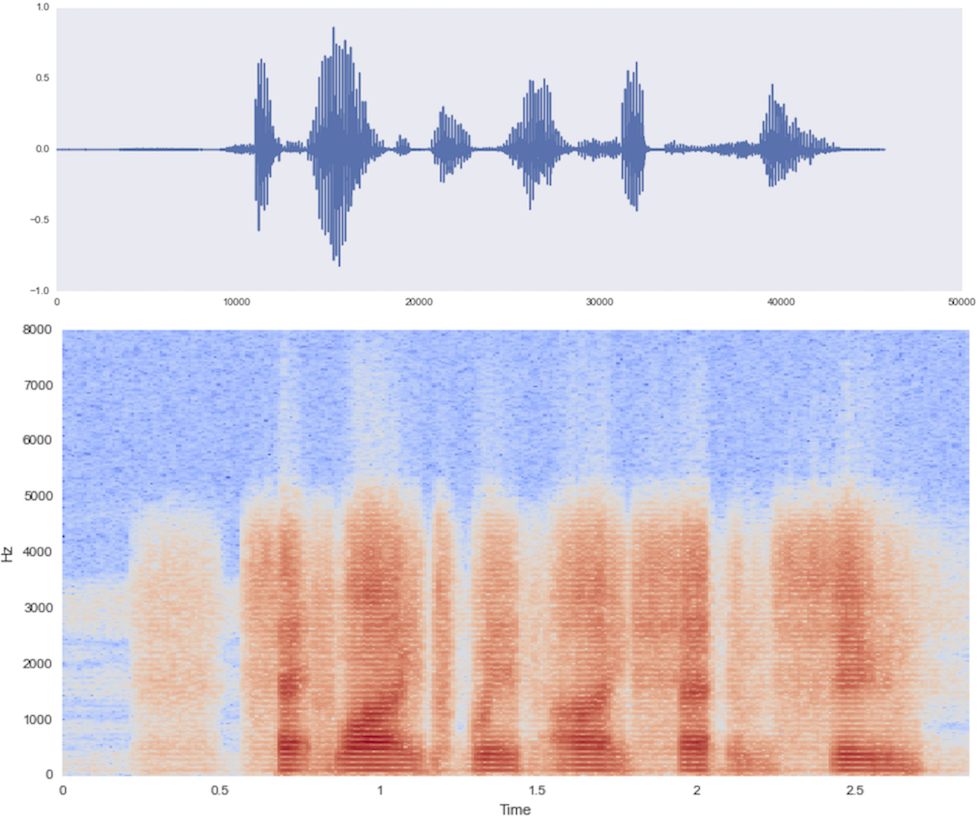}
\caption[]%
{{\small Le Cornu \emph{et al.} \cite{cornu2017}}}    
\end{subfigure}
\hfill
\begin{subfigure}[b]{0.475\textwidth}   
\centering 
\includegraphics[width=\textwidth]{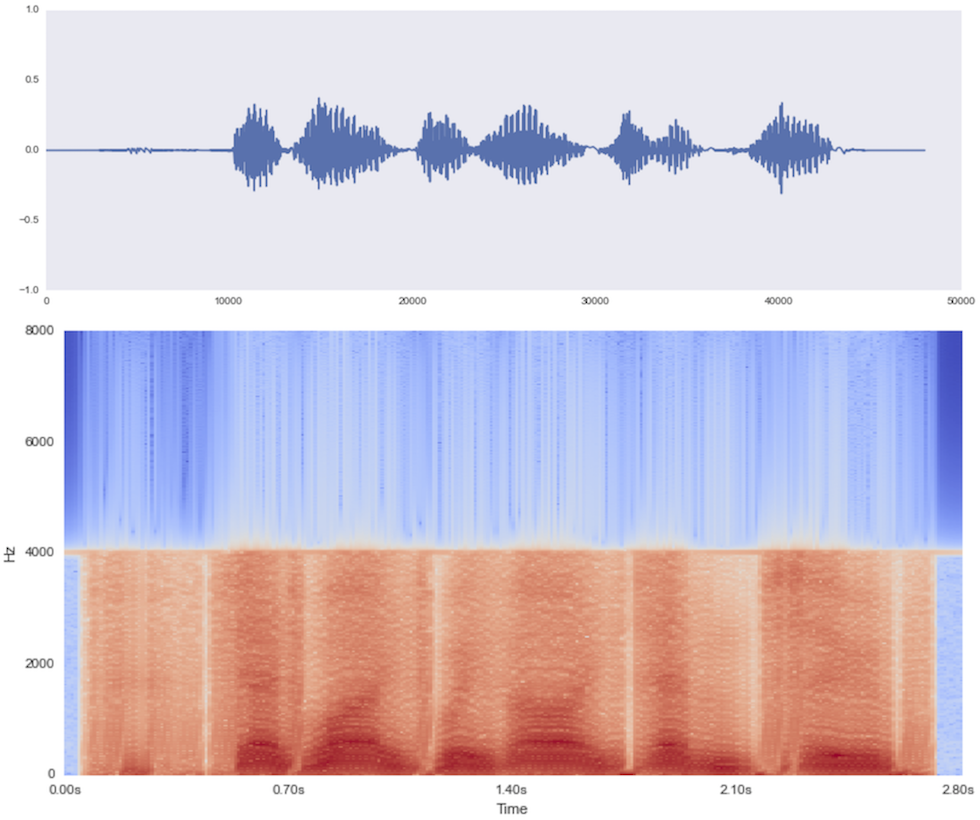}
\caption[]%
{{\small Vid2speech \cite{vid2speech}}}    
\end{subfigure}
\caption[]
{\small Visualization of original vs. reconstructed speech results on one GRID speaker $S3$ utterance. Original waveform and spectrogram are shown in (a). (b) shows the results of our reconstruction. The results of \cite{cornu2017} and \cite{vid2speech} are depicted in (c) and (d), respectively. Best viewed on a color display.} 
\label{fig:result_viz}
\end{figure*}

\subsection{Generating a waveform}
\label{ssec:waveform}
We consider several methods for generating a waveform from our model's predicted mel and linear scale sound features. The first is the spectrogram synthesis approach used by \cite{tacotron,nsynth} in which the Griffin-Lim algorithm is used to reconstruct the phase of the predicted linear-scale spectrogram. Inverse STFT is then used to convert the complex spectrogram back into waveform. We found that the result is intelligible and smooth, albeit somewhat unnatural and robotic sounding.

Therefore, we also consider an example-based synthesis method similar to the one used by \cite{owens2015visually}, in which we replace predicted sound features with their closest exemplar in the training set. We search for the nearest neighbor to both mel-scale and linear-scale predicted features, as measured by $L_2$ distance, and replace it with the neighbor's corresponding linear-scale spectrogram feature. The full spectrogram is then converted into waveform using the procedure described above. We find that in most cases, mel-scale gives better results, which are more natural-sounding, but less smooth than using the predicted linear spectrogram itself.

\section{Model details}
\label{sec:details}
We use the method of \cite{sun2013deep}, with the code provided from their website, to detect facial features. The speaker's face is cropped to $160 \times 128$ pixels. Using $K=9$ frames as input worked best. This results in an input volume of size $160 \times 128 \times 9$ scalars, from which we subtract the mean. We use the method of \cite{liu2009beyond} with python wrapper provided by \cite{pathak2016learning} to compute an optical flow vector $(u,v)$ for every image pixel. Optical flow is not normalized as its mean is approximately zero, and its std is in the range of $0.5-1.5$ pixels. \cite{jampani2016video,pathak2016learning}. 

We use the code provided by \cite{nsynth} to compute log magnitude of both linear and mel-scale spectrograms which are peak normalized to be between $0$ and $1$. For videos with a frame rate of $25$ FPS we downsample the original audio to $16$ kHz and use $40$ ms windows with $10$ ms frame shift. For videos with a frame rate of $29.97$ FPS we downsample the original audio to $14985$ Hz and use $33.3$ ms windows with $8.3$ ms frame shift.

Our network implementation is based on the \emph{Keras} library \cite{chollet2015keras} built on top of \emph{TensorFlow} \cite{tensorflow2015-whitepaper}. Network weights are initialized using the initialization procedure suggested by He \emph{et al.} \cite{He_2015_ICCV}. Before each activation layer Batch Normalization \cite{ioffe2015batch} is performed. We use Leaky ReLU \cite{maas2013rectifier} as the non-linear activation function in all layers but the last two, in which we use the hyperbolic tangent (tanh) function. Adam optimizer \cite{kingma2014adam} is used with an initial learning rate of $0.001$, which is reduced several times during training. Dropout \cite{srivastava2014dropout} is used to prevent overfitting, with a rate of $0.25$ after convolutional layers and $0.5$ after fully connected ones. We use mini-batches of $16$ training samples each and stop training when the validation loss stops decreasing (around $100$ epochs). 
The network is trained with backpropagation using mean squared error ($L_2$) loss for both decoder and post-processing net outputs, which have equal weights. To improve the temporal smoothness of the output, after generating spectrogram coefficients for $T$ consecutive frames ($1 ... T$), we move one step forward and do the same for ($2 ... T{+}1$), which creates an overlap of $T-1$ frames, thus creating exactly $T$ predictions for each frame. We then calculate a weighted average over the predictions for a given frame using a Gaussian.

\section{Experiments}
\label{sec:experiments}

\paragraph{Datasets}
Previous works performed experiments with the GRID audiovisual sentence corpus \cite{gridcorpus}, a large dataset of audio and video (facial) recordings of $1000$ sentences spoken by $34$ talkers ($18$ male, $16$ female). Each sentence consists of a six word sequence of the form shown in Table \ref{tb:grid}, e.g. ``Place green at H $7$ now". Although this dataset contains a fair volume of high quality speech videos, it has several limitations, the most important being its extremely small vocabulary. In order to compare our method with previous ones, we too perform experiments on this dataset.

In order to better demonstrate the capability of our model, we also perform experiments on the TCD-TIMIT dataset \cite{tcd-timit}. This dataset consists of $60$ volunteer speakers with around $200$ videos each, as well as three \emph{lipspeakers}, people specially trained to speak in a way that helps the deaf understand their visual speech. The speakers are recorded saying various sentences from the TIMIT dataset \cite{timit}, and are recorded using both front-facing and $30$ degree cameras.

\paragraph{Evaluation}
We evaluated the quality and intelligibility of the reconstructed speech using four well-known objective scores: STOI \cite{stoi} and ESTOI \cite{jensen2016algorithm} for estimating the intelligibility of the reconstructed speech and automatic mean opinion score (MOS) tests PESQ \cite{pesq} and VisQOL \cite{visqol}, which indicate the quality of the speech. While all objective scores have their faults, we found that these metrics correlate relatively well with our perceived audio quality, as well as with our model's loss function. However, we strongly encourage readers to watch and listen to the supplementary video on our project web page \footnote{Examples of reconstructed speech can be found at \\ \url{http://www.vision.huji.ac.il/vid2speech}} which conclusively demonstrates the superiority of our approach.

\subsection{Sound prediction tasks}
\label{ssec:tasks}

\begin{table}
\centering
\begin{tabular}{@{ }c@{\hspace{3mm}}@{}c@{\hspace{3mm}}@{}c@{\hspace{2mm}}@{}c@{\hspace{2mm}}@{}c@{\hspace{3mm}}@{}c@{ }}
\toprule[1.5pt]
\bf Command&\bf Color&\bf Preposition&\bf Letter&\bf Digit &\bf Adverb	\\
\midrule
bin 	&  blue		& at  	&  A-Z		& 0-9	& again      \\
lay 	&  green 	& by  	&  minus W 	&   	& now      	 \\
place	&  red		& in 	&			& 		& please 	 \\
set		&  white	& with 	& 			& 		& soon		 \\
\bottomrule[1.5pt] \\
\end{tabular}
\caption{GRID sentence grammar.}
\label{tb:grid}
\end{table}

\paragraph*{GRID}
For this task we trained our model on a random $80/20$ train/test split of the $1000$ videos of speakers $S1-3$ (male) and $S4$ (female), and made sure that all $51$ GRID words were represented in each set. The resulting representation vectors were converted back into waveform using the aforementioned mel-scale spectrogram example-based synthesis (\emph{Mel-synth}) and predicted linear spectrogram synthesis (\emph{Lin-synth}).

\begin{table}
\centering
\begin{tabular}{lcccc}
\toprule[1.5pt]
\bf  & \bf STOI & \bf ESTOI & \bf PESQ & \bf ViSQOL\\
\midrule
\bf GRID S1 & \\
\midrule
Mel-synth 	&  $0.442$ & $0.26$ & $1.782$ & $3.108$\\
Lin-synth	& $\bf0.475$ & $\bf0.263$ & $\bf1.952$ & $\bf3.324$\\
\midrule
\bf GRID S2 & \\
\midrule
Mel-synth 	&  $0.631$ & $0.434$ & $2.107$ & $3.07$\\
Lin-synth	& $\bf0.667$ & $\bf0.462$ & $\bf2.136$ & $\bf3.316$\\
\midrule
\bf GRID S3 & \\
\midrule
Mel-synth 	&  $0.666$ & $\bf0.398$ & $\bf1.974$ & $3.164$\\
Lin-synth	& $\bf0.68$	& $0.354$ & $1.904$ & $\bf3.349$\\
\midrule
\bf GRID S4 & \\
\midrule
Mel-synth 	&  $0.644$ & $0.429$ & $1.809$ & $3.092$\\
Lin-synth	& $\bf0.7$	& $\bf0.462$ & $\bf1.922$ & $\bf3.3$\\
\bottomrule[1.5pt] \\
\end{tabular}
\caption{Objective measurements of reconstructed speech for model trained on GRID speakers $S1-4$. Mel-scale spectrogram example-based synthesis (\emph{Mel-synth}) and predicted linear spectrogram synthesis (\emph{Lin-synth}) were used to generate waveform from the predicted audio representations.}.
\label{tb:s3_results}
\end{table}

\begin{table}
\centering
\begin{tabular}{lcccc}
\toprule[1.5pt]
\bf  & \bf STOI & \bf ESTOI & \bf PESQ & \bf ViSQOL\\
\midrule
\bf GRID S3 & \\
\midrule
Vid2speech \cite{vid2speech} & $0.638$ & $-$ & $1.875$ & $3.13$ \\
Cornu \etal\cite{cornu2017} 	& $-$ & $\bf0.437$ & $\bf2.055$ & $-$\\
Ours	& $\bf0.68$	& $0.398$ & $1.974$ & $\bf3.349$\\
\midrule
\bf GRID S4 & \\
\midrule
Vid2speech \cite{vid2speech} & $0.584$ & $-$ & $1.19$ & $2.691$ \\
Cornu \etal\cite{cornu2017} 	& $-$ & $0.434$	&  $1.686$ & $-$\\
Ours	& $\bf0.7$ & $\bf0.462$ & $\bf1.922$ & $\bf3.3$\\
\bottomrule[1.5pt] \\
\end{tabular}
\caption{Comparison to \cite{vid2speech} and \cite{cornu2017} using objective measurements.}
\label{tb:lecornu_comparison}
\end{table}

Table \ref{tb:s3_results} shows the results of reconstructing the speech of $S1-4$ and Table \ref{tb:lecornu_comparison} shows a comparison of our best results to the best results of \cite{cornu2017}. Figure \ref{fig:result_viz} shows a visualization of our results on one $S3$ video, and compares it to the original, and the results of \cite{cornu2017,vid2speech}.

\paragraph*{TCD-TIMIT}
For this task we trained our model on a random $90/10$ train/test split of the $254$ videos of Lipspeakers $1-3$ (female). This results in less than $25$ minutes of video used for training, which only around $60\%$ of the amount of data used in the previous task. In this task, many of the words in the test set do not appear in the training set. We would like for our model to learn to reconstruct these words based on the recognition of the combinations of short visual segments which comprise the words. Here too, the resulting audio vectors were converted back into waveform using mel-scale spectrogram example-based synthesis (\emph{Mel-synth}) and predicted linear spectrogram synthesis (\emph{Lin-synth}).

Table \ref{tb:lipspkr3_results} holds the results of this difficult task. The reconstructed speech is natural-sounding, albeit not entirely intelligible. Given the limited amount of training data, we believe our results are promising enough to indicate that fully intelligible reconstructed speech from unconstrained dictionaries is a feasible task.

\begin{table}
\centering
\begin{tabular}{lcccc}
\toprule[1.5pt]
\bf  & \bf STOI & \bf ESTOI & \bf PESQ & \bf ViSQOL \\
\midrule
\bf Lipspeaker 1 & \\
\midrule
Mel-synth 	& $0.274$ & $0.116$ & $1.309$ & $2.533$\\
Lin-synth	& $\bf0.318$ & $\bf0.171$ & $\bf1.384$ & $\bf2.943$\\
\midrule
\bf Lipspeaker 2 & \\
\midrule
Mel-synth 	& $0.307$ & $0.123$ & $1.08$ & $2.431$\\
Lin-synth	& $\bf0.357$ & $\bf0.157$ & $\bf1.158$ & $\bf2.931$\\
\midrule
\bf Lipspeaker 3 & \\
\midrule
Mel-synth 	&  $0.565$ & $0.373$ & $1.484$ & $2.834$\\
Lin-synth	& $\bf0.63$	& $\bf0.447$ & $\bf1.612$ & $\bf3.201$\\
\bottomrule[1.5pt] \\
\end{tabular}
\caption{Objective measurements of reconstructed speech for model trained on TCD-TIMIT Lipspeakers 1-3. Many words in the testing set were not present in the training set, which results in significantly worse results than those on GRID dataset. The Lipspeaker 3 model performs best, by a large margin.}.
\label{tb:lipspkr3_results}
\end{table}

\subsection{Ablation analysis}
\label{ssec:ablation}
We conducted a few ablation studies on GRID speaker $S3$ to understand the key components in our model. We compare our full model with ($i$) a model using only an optical flow stream; ($ii$) a model using only a pixel stream; ($iii$) a model with no post-processing network which outputs mel-scale spectrogram. Table \ref{tb:ablation} shows the results of this analysis. Our analysis shows that pixel intensities provide most of the information needed for reconstructing speech, while adding optical flow and a post-processing network give slightly better results.

\begin{table}
\centering
\begin{tabular}{lccc}
\toprule[1.5pt]
\bf  & \bf STOI & \bf PESQ & \bf ViSQOL \\
\midrule
Optical flow only	&  $0.381$ & $1.478$ & $2.786$\\
Pixels only	& $0.67$	&  $1.949$ & $3.179$\\
Pixels + optical flow 	&  $0.665$ & $1.921$ & $3.173$\\
\bf Pixels + OF + postnet	& $\bf0.68$	&  $\bf1.974$ & $\bf3.349$\\
\bottomrule[1.5pt] \\
\end{tabular}
\caption{Results of ablation analysis on GRID speaker $S3$. Pixels provide most of the information needed, adding optical flow and post-processing network gives slightly better results}.
\label{tb:ablation}
\end{table}

\section{Concluding remarks}
\label{sec:conclusion}

A two-tower CNN-based model is proposed for reconstructing intelligible and natural-sounding speech from silent video frames of a speaking person. The model is trained end-to-end with a post-processing network which fuses together multiple CNN outputs in order to obtain a longer range speech representation. We have shown that the proposed model obtains significantly higher quality reconstructions than previous works, and even shows promise towards reconstructing speech from an unconstrained dictionary.

The work described in this paper can be improved upon by increasing intelligibility of speech reconstruction from an unconstrained dictionary, and extending an existing model to unknown speakers. It can also be used as a basis for various speech related tasks such as speaker separation and enhancement. 

We plan to use the visually reconstructed speech in order to enhance speech recorded in a noisy environment, and to separate mixed speech in scenarios like the ``cocktail party'' where the face of the speaking person is visible \cite{gabbay2017seeing}.

~\\
\noindent
{\bf Acknowledgment.} This research was supported by Israel Science Foundation, by DFG, and by Intel ICRI-CI.

{\small
\bibliographystyle{ieee}
\bibliography{egbib}
}

\end{document}